\documentclass[aps,prx,nofootinbib,floatfix,twocolumn,longbibliography,superscriptaddress]{revtex4-2}

\usepackage[british]{babel}
\usepackage{amsmath,amsfonts,amssymb,amsthm}
\usepackage{booktabs}
\usepackage{graphicx}
\usepackage{subcaption}
\usepackage{xcolor}
\usepackage{hyperref}
\usepackage{braket}
\usepackage{float}

\hypersetup{hidelinks}

\newcommand{\MI}{\mathrm{MI}}

\newcommand{\beq}{\begin{equation}}
\newcommand{\eeq}{\end{equation}}

\begin{document}

\title{
Do Quantum Models Scale Like LLMs?
% Neural Scaling in a Quantum Critical Model
}

\author{David S. Berman}
\affiliation{Centre for Theoretical Physics, Queen Mary University of London,
Mile End Road, London E1 4NS, United Kingdom}

\author{Ying-Jer Kao}
\affiliation{Department of Physics, National Taiwan University,
Taipei 10607, Taiwan}

\author{Roger G. Melko}
\affiliation{Department of Physics and Astronomy, University of Waterloo,
Waterloo, Ontario N2L 3G1, Canada}
\affiliation{Perimeter Institute for Theoretical Physics,
Waterloo, Ontario N2L 2Y5, Canada}

\author{Alexander G. Stapleton}
\affiliation{Centre for Theoretical Physics, Queen Mary University of London,
Mile End Road, London E1 4NS, United Kingdom}

\begin{abstract}
In this work, we study the neural scaling laws of RydbergGPT, an autoregressive transformer
model trained on qubit projective measurement data gathered from interacting Rydberg atom arrays. 
%We examine how the
%neural scaling laws governing model training depend on the quantum properties of the system.
%The laser detuning provides a control parameter allowing the training data to be partitioned,
%meaning one may study the neural scaling laws as a function of detuning. 
The quantum system is known to exhibit a finite-size remnant of a critical point as the laser detuning parameter is varied.
We find that near the critical point the transformer loss as a function of training dataset size is
well described by a power-law with a loss floor correction. 
However, away from criticality the quality of the power-law description is substantially reduced. 
%This suggests that stable neural scaling depends not only on model architecture, but also features of the training distribution.
We then compare the statistical structure of
both Rydberg measurements and natural-language corpora using an entropy-normalised, finite
sample corrected mutual information `two-point' function. We find that near-critical statistics
of the two point functions are closest to those observed in natural-language, whilst other qubit
configurations far from the critical point have two-point functions that decay more rapidly.
This supports the hypothesis that multi-scale dependence contributes to stable neural scaling, and that scaling behaviour should be viewed as a property of the model–data pair.
\end{abstract}

\maketitle

\section{Introduction}
\label{sec:introduction}

In recent years, transformer models \cite{vaswani2017attention} have become pervasive across a wide range of applications. Of these, arguably the most ubiquitous use of transformers is in models which produce and process natural language. Such networks often have a vast number of parameters, and are thus almost universally referred to as \textit{large language models} (LLMs). Undoubtedly, LLMs represent one of the most significant technological advances of the past century; however, as with many modern and competitive machine-learning models, their training is computationally expensive, time-consuming, and associated with substantial environmental costs. Understanding how large models scale is thus of central importance. By relating the computational resources required for training to the resulting predictive loss, an effective neural scaling law should enable the forecasting resources, allow one to make cheap comparisons between learning regimes, and assess how efficiently a model captures the statistical structure of its training data. 

%In common parlance, neural scaling laws typically refer to results associated with models trained on various domains of \textit{natural language}. Since this note primarily concerns quantum data, we make a distinction between scaling laws for LLMs, or \textit{natural language neural scaling laws}, and \textit{neural scaling laws} for general machine learning models.
In this paper, we ask whether transformers trained on quantum measurement data exhibit similar neural scaling laws, and therefore whether their performance could reliably be forecast as increasingly large datasets from quantum devices become available for training.

In general, common `neural scaling laws' are actually almost exclusively \textit{natural language} neural scaling laws. Of these, the two most famous are \textit{Hoffmann scaling} \cite{hoffmann2022trainingcomputeoptimallargelanguage} and \textit{Kaplan scaling} \cite{kaplan2020scalinglaws}. As the number of model parameters and the training tokens (i.e. subwords) increase, the predictive loss is shown to follow a remarkably regular and predictable trend. The success of these frameworks motivates a broader question: 
is scaling behaviour a distinctive property of natural language, or does it reflect a more general feature of learning from structured probability distributions with a transformer-based architecture?

Whilst some works have endeavoured to derive scaling properties of natural language from their statistics, for example \cite{cagnetta2026derivingneuralscalinglaws}, the generalisation of similar statistical measures to broader contexts remains remarkably under-studied. 
Recent theoretical works studying datasets motivated by physical systems suggest that neural scaling can depend crucially on the statistical structure of the training distribution \cite{Bahri,Barkeshli,Peraza}.

In this work, we investigate this question using RydbergGPT \cite{Fitzek_2025}, an open-source transformer model trained on synthetic measurement samples generated by quantum Monte Carlo simulations of a neutral Rydberg atom array. 
%Specifically, we ask does \emph{quantum data scale like natural language?} The Rydberg model provides an excellent test bed for this investigation owing to the physical interpretability of the parameters governing the data-generating distribution. These parameters allow scaling behaviour to be examined systematically, and predictive loss to be compared directly across different physical regimes.
%RydbergGPT learns the conditional probability distribution of the occupation of Rydberg lattice sites given previous spatially local sites. 

Rydberg atom arrays are programmable quantum computing devices, that combine flexible control of qubit interactions with single-atom resolved preparation and measurement \cite{Ebadi}.

They are
composed of atoms in which one valence electron is excited to a state with a very large principal quantum number \(n\). In such states the electron is only weakly bound and occupies an orbital whose radius scales approximately as \(n^2\), giving rise to exaggerated atomic properties. In particular, the electric dipole moment and polarisability become very large, meaning that two Rydberg atoms can interact strongly even when separated by relatively large distances; see \cite{saffman2010quantum,browaeys2020manybody} for reviews of the platform and its use in quantum simulation.

The \textit{Rydberg model} treats each atom as an effective two-level qubit system consisting of an electronic ground state \(\ket{g}\) and a Rydberg state \(\ket{r}\). Following the notation of RydbergGPT \cite{Fitzek_2025} (and the standard square-lattice Rydberg-array Hamiltonians used in studies of density-wave order and quantum phase transitions \cite{samajdar2020complex,kalinowski2022bulk}), for a square $L\times L$ array of atoms at position vectors \(\{r_i\}_{i=1}^N\), where $N=L^2$, the Hamiltonian may be written as
\beq
\hat{H}
=
\sum_{i<j}
\frac{C_6}{\|r_i-r_j\|^6}
\hat{n}_i\hat{n}_j
-
\delta \sum_{i=1}^N \hat{n}_i
-
\frac{\Omega}{2}\sum_{i=1}^N \hat{\sigma}_i^x ,
\eeq
where
\beq
\hat{\sigma}_i^x
=
\ket{g}_i\bra{r}_i+\ket{r}_i\bra{g}_i,
\quad
\hat{n}_i
=
\frac{1}{2}\left(\hat{\sigma}_i+1\right)
=
\ket{r}_i\bra{r}_i,
\nonumber
\eeq
\beq
\hat{\sigma}_i
=
\ket{r}_i\bra{r}_i-\ket{g}_i\bra{g}_i .
\eeq
The interaction strength may equivalently be parameterised by the blockade radius \(R_b\) and lattice spacing \(a\),
\beq
C_6 = \Omega \left(\frac{R_b}{a}\right)^6,
\qquad
V_{ij}
=
\frac{a^6}{\|r_i-r_j\|^6}.
\eeq

The parameter \(\Omega\) sets the coherent drive between \(\ket{g}\) and \(\ket{r}\), while the laser detuning \(\delta\) biases the energy cost of creating a Rydberg excitation through the term \(-\delta\sum_i \hat n_i\). The ratio \(R_b/a\) fixes the effective interaction scale through \(C_6\), the matrix \(V\) encodes the lattice geometry via the separations \(\|r_i-r_j\|\).
The final physical parameter is the Rydberg blockade radius $R_b$ which penalises simultaneous excitation of nearby atoms. 
In addition, in the synthetic RydbergGPT dataset 
\cite{RGPTdata}
produced by world-line QMC,
an effective inverse temperature 
\(\beta\Omega\) 
is included in order to produce thermal ensembles.
%controls the effective inverse temperature of the thermal ensemble used to generate measurements. 
%Notice the inverse temperature \(\beta\) does not appear as a term in the Hamiltonian \(\hat H\); rather, it determines which thermal state is sampled through \(\rho_\beta = e^{-\beta \hat H}/\mathrm{Tr}(e^{-\beta \hat H})\). 
%In the following sections, the dimensionless quantity \(\beta\Omega\) is supplied as part of the conditioning input, so changing \(\beta\Omega\) changes the measurement distribution without changing the operator form of \(\hat H\). 

The competition between laser driving, detuning, interactions and temperature produces structured many-body measurement distributions.

\subsection{RydbergGPT}
\label{sub:rydberggpt}

RydbergGPT is an autoregressive transformer model, parameterised by weights $\theta$, designed to learn measurement distributions of Rydberg systems \cite{Fitzek_2025}. Architecturally, it inherits the transformer/autoregressive modelling structure introduced in \cite{vaswani2017attention}, but conditions generation on Hamiltonian data rather than natural language context. In the notation of the previous section, the physically controllable parameters of the model are encoded by
\beq
x
=
\left(
\Omega,\,
\delta/\Omega,\,
R_b/a,\,
V,\,
\beta\Omega
\right),
\eeq
and the target sequence is a binary occupation-basis measurement
\beq
\sigma=\{\sigma_1,\sigma_2,\ldots,\sigma_N\}.
\eeq
The model learns the conditional probabilities
\beq
p_\theta(\sigma;x)
=
\prod_{i=1}^{N}
p_\theta\!\left(\sigma_i \mid \sigma_{<i};x\right),
\eeq
where \(\theta\) denotes the trainable neural-network parameters.

In this machine-learning formulation, \(x\) is the condition supplied to the encoder: changing $x$ induces a change in the physical properties of the distribution that the decoder learns. The sequence variables \(\sigma_i\) play the role of tokens, while \(\theta\) controls the expressive capacity of the transformer used to approximate the Rydberg array measurement distribution. Each configuration is canonically tokenised such that $\sigma_i$ correspond to bits in $\{0,1\}$, representing $\ket{g}$ and $\ket{r}$ respectively.

\begin{figure}[H]
  \centering
  \includegraphics[width=0.98\linewidth]{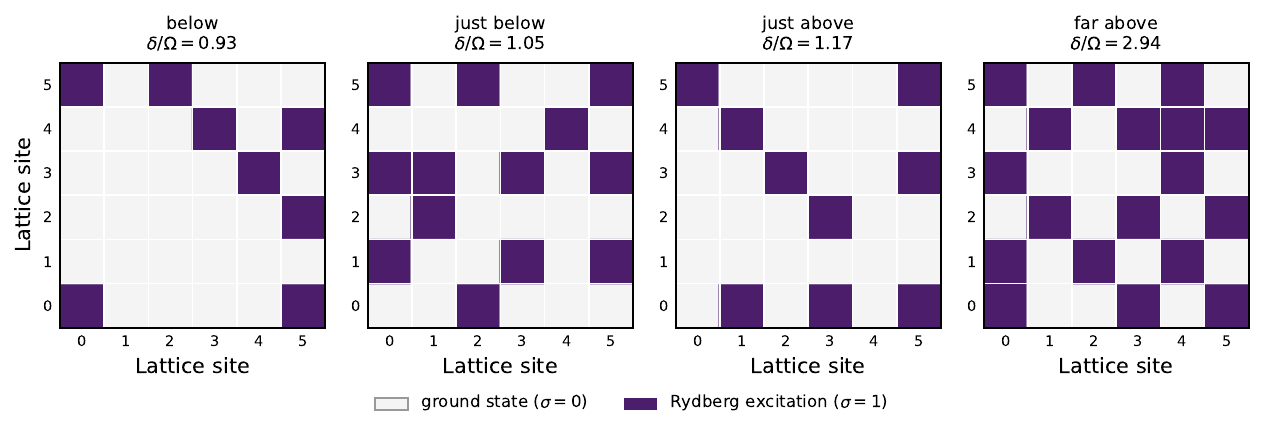}
  \caption{Representative randomly selected occupation-basis configurations from an open $6\times6$ Rydberg array at the four labelled detunings \cite{RGPTdata}. Light sites are in the ground state and dark sites represent Rydberg excitations. The samples were selected with a fixed pseudorandom seed from the same data archive used for the data-scaling and mutual-information analyses. They are individual measurements and are shown for visual orientation only.}
  \label{fig:detuning-reference-samples}
\end{figure}

For visual reference, Figure~\ref{fig:detuning-reference-samples} shows four single occupation-basis measurements from the same ensembles. The samples are selected reproducibly without conditioning on their spatial pattern. They illustrate the binary configurations presented to the model, rather than an ensemble-averaged order parameter or correlation function.

As was alluded to earlier, the physical interpretability of $x$ makes RydbergGPT a useful test case for scaling behaviour beyond natural language. In language, loss scaling is usually interpreted in terms of the number of text tokens, the number of model parameters (which vicariously encode the compute budget), and the acceptable loss value. Conversely, in the Rydberg setting, the amount of data is only one part of the story. Samples at different detunings possess very different internal structure, ranging from weakly dependent random fields in the very high temperature regime to more constrained spatial patterns nearer criticality, for example. 

One way to consider this is through effective dataset size. In a maximally degenerate dataset the number of data samples to fully specify the target distribution is simply unity, whereas in the random noise limit, the number of data samples to specify the distribution becomes infinite. Of course, in large limits the distribution of text is effectively not random (for example Zipf's law applies) in the same way that quantum systems at criticality are not random. Schematically, one may write the effective size of the dataset $D_\text{eff}$ as a scaling of a Zipfian text dataset, namely $D_\text{eff} = \Gamma D_\text{text}$. In the case that $\Gamma > 1$, the training dataset in some sense encodes more information than an equivalently sized corpus for natural language. Conversely, if $\Gamma < 1$, then the corpus encodes less information than an equivalently sized natural language corpus. 

A study of the dependence on effective dataset size is crucial since it asks whether the empirical scaling principles developed for language models are properties of text itself\footnote{Natural language exhibits incredible structure: Zipf's law \cite{KingsleyZipf1932} holds almost universally, for example \cite{berman2026pathnaturallanguagetokenisation}.} or of a broader class of structured distributions.
If similar scaling behaviour appears in a controlled physical setting, it becomes possible to relate model performance not only to dataset size and parameter count, but also to interpretable features of the underlying quantum system. A natural way to make this dependence structure precise is through mutual information. The information-theoretic foundation for this is entropy \cite{shannon1948mathematical}; mutual-information functions have long been used to diagnose statistical dependence in natural language \cite{li1989mutualinformation,ebeling1994entropy}. The connection between mutual-information decay, the structure of natural language, and statistical physics of states at criticality was made explicit by Lin and Tegmark \cite{lin2017criticalbehavior}; although more recent work has also used mutual information as a scaling object for long-context language modelling \cite{chen2025l2m}. These references motivate treating raw token or sample count as only one coordinate of the learning problem.

\subsection{A General Review of Hoffmann Scaling}
\label{sub:chinchilla}

Whilst there are many known scaling laws which empirically provide a good fit to experimental data, arguably one of the most ubiquitous is \textit{Hoffmann scaling}\footnote{An equally prevalent scaling law is known as \textit{Kaplan scaling}, initially presented in \cite{kaplan2020scalinglaws}.}. Hoffmann scaling laws were developed for transformer language models trained under a fixed compute budget \cite{hoffmann2022trainingcomputeoptimallargelanguage}, building on the broader empirical scaling-law programme of \cite{kaplan2020scalinglaws}. The main qualitative lesson is that compute-optimal training should not only increase the number of parameters; it should also increase the number of training tokens. A model that is too large for its dataset is under-trained, while a model that is too small may fail to use available data efficiently. In the usual language-modelling notation, the expected loss is treated as a function of parameter count $P$ and token count $D$, for example through an empirical decomposition of the form
\beq
  \label{eqn:chinchilla_vanilla}
  \mathcal{L}(P,D)
  \approx
  \mathcal{L}_{\infty}
  + A P^{-\alpha}
  + B D^{-\gamma},
\eeq
where $\mathcal{L}_{\infty}$ is an irreducible loss floor and the two power-law terms describe limitations from finite model size and finite data. Chinchilla scaling then asks how $P$ and $D$ should co-vary when the training compute budget is held fixed.

Since quantum data does not exhibit a universal structure (like that in natural language which gives rise to Zipf's law), a useful empirical model for the RydbergGPT loss arises from promoting the scaling constants in Equation \eqref{eqn:chinchilla_vanilla} to functions of the physical parameters, i.e.
\beq
\label{eqn:rydberg_loss}
  \mathcal{L}_{\mathrm{Ryd}}(P,D;x)
  \approx
  \mathcal{L}_{\infty}(x)
  + A(x) P^{-\alpha(x)}
  + B(x) D(x)^{-\gamma(x)}.
\eeq
Following the spirit of \cite{kaplan2020scalinglaws,hoffmann2022trainingcomputeoptimallargelanguage}, this form is not assumed to be exact, but rather to separate three possible causes of improved validation loss: increased model capacity, increased raw sample count, and changes in the physical complexity of the data distribution. In the remainder of this work, we hold all parameters in \(x\) fixed except the laser detuning \(\delta\), and use that as the physical control parameter.

\section{Fixed-Parameter Data Scaling of RydbergGPT}
\label{sec:fixed_param_scaling_rydberg}

We first consider data scaling at fixed model size. Let $P_0$ denote the fixed number of trainable parameters and let $D$ denote the number of training configurations, or equivalently the number of lattice-token sequences used during training. The dataset size is swept at fixed $L=6$, $R_b/a=1.15$, and $\beta\Omega=8$ for the completed detuning set
\begin{equation}
\delta/\Omega\in\{-0.36,0.93,1.05,1.17,1.52,2.94\}.
\label{eq:completed-detuning-sweep}
\end{equation}
The detunings $1.05$ and $1.17$ bracket the known quantum critical regime of this model.  

Note, due to the small lattice sizes considered here, 
we only have access to the finite-size remnant of the
true quantum phase transition.
We nonetheless refer to this of delta as the {\em critical point} in the below discussion.
%critical finite-array crossover around the infinite lattice critical point. Due to finite-size effects, these are treated as nearby reference points rather than as values on opposite sides of a sharply defined thermodynamic transition. Moreover, the transverse drive introduces quantum fluctuations which round what would be a phase transition into a quantum crossover.
%For the sake of clarity and intuition, we will refer to the critical crossover as \textit{near criticality} or the \textit{critical crossover}. 
As was found in \cite{kalinowski2022bulk} which employed quantum Monte Carlo simulations to model the near-critical region, $\delta/\Omega\simeq1.1$ may be used as a reference value for the critical point.

For each detuning, the validation loss is measured as
\begin{equation}
\widehat{\mathcal{L}}_{\mathrm{val}}(D;\delta)
\sim-\frac{1}{M_{\mathrm{val}}}
\sum_{m=1}^{M_{\mathrm{val}}}\sum_{t=1}^{T_m}
\log p_{P_0,D}\!\left(\sigma_t^{(m)}
\mid\sigma_{<t}^{(m)},\delta\right),
\label{eq:validation-loss}
\end{equation}
where $\sigma_t^{(m)}$ is the $t$-th site token in the \(m\)-th held-out draw from the Rydberg ensemble, and $p_{P_0,D}$ is the trained RydbergGPT predictive distribution at fixed parameter count $P_0$. For each data fraction and seed, we take the minimum logged training or validation loss, average over ten seeds, and fit the mean curve. The data fractions range from $10\%$ to $100\%$ of the available training data.

At fixed model size, we fit the aggregate curves to the empirical form
\begin{equation}
\mathcal L(f;\delta)=\mathcal L_\infty(\delta)
+B_D(\delta) D(f)^{-\gamma_D(\delta)},
\label{eq:data-scaling}
\end{equation}
where $f$ is the training-data percentage, $\mathcal L_\infty$ is the fitted loss floor, and $\gamma_D$ is the data-scaling exponent.

\begin{figure}[H]
  \centering  \includegraphics[width=0.98\linewidth]{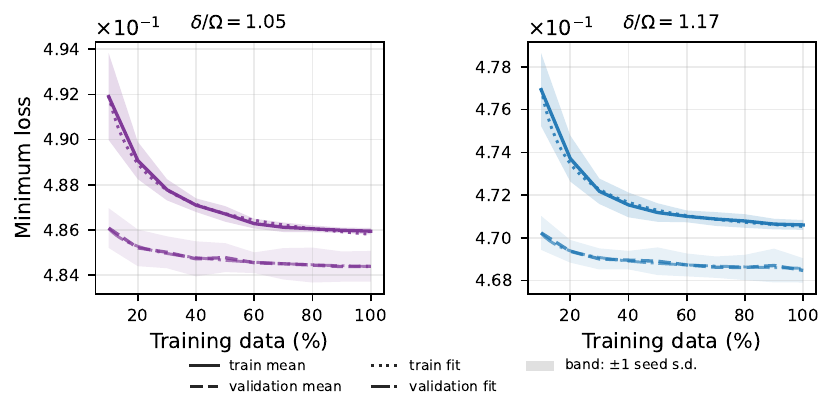}
  \caption{Fixed-model RydbergGPT data-scaling curves on the two sides of the critical detuning point. The system parameters are $L=6$, $R_b/a=1.15$ and $\beta\Omega=8$. Solid and dashed curves denote the seed-averaged training and validation losses, respectively; shaded regions show one standard deviation across seeds. Dotted and dash-dotted curves are fits to Equation~\eqref{eq:data-scaling}. The scientific-notation multiplier is shown once per vertical axis.}
  \label{fig:detuning-static-losses}
\end{figure}

Figure~\ref{fig:detuning-static-losses} compares the two completed detunings that bracket the critical point. Both the training and validation curves decrease systematically with training-set fraction.

\begin{table}[H]
  \centering
  \small
  \begin{tabular}{cccccc}
    \toprule
    $\delta/\Omega$ & series & $B_D$ & $\gamma_D$ &
    $\mathcal L_\infty$ & $R^2$ \\
    \midrule
    1.05 & train & 0.0409 & 0.736 & 0.4844 & 0.998 \\
    1.05 & validation & 0.00980 & 0.656 & 0.4839 & 0.989 \\
    1.17 & train & 0.0647 & 0.950 & 0.4697 & 0.997 \\
    1.17 & validation & 0.0121 & 0.791 & 0.4683 & 0.983 \\
    \bottomrule
  \end{tabular}
  \caption{Aggregate fits to Equation~\eqref{eq:data-scaling}. Each fit uses ten data fractions. The validation loss floor is lower at $\delta/\Omega=1.17$ than at $1.05$, while the validation exponent remains of order unity on both sides of the critical point.}
  \label{tab:detuning-data-scaling}
\end{table}

The fitted validation floor decreases by $3.2\%$ between $\delta/\Omega=1.05$ and $1.17$, from $0.4839$ to $0.4683$. The corresponding exponent increases from $0.656$ to $0.791$. The high fit qualities in Table~\ref{tab:detuning-data-scaling} show that the observed aggregate curves are adequately represented by the three-parameter form within this narrow interval. They do not, by themselves, identify a universal exponent or a discontinuity.

\begin{figure}[H]
  \centering
  \includegraphics[width=0.90\linewidth]{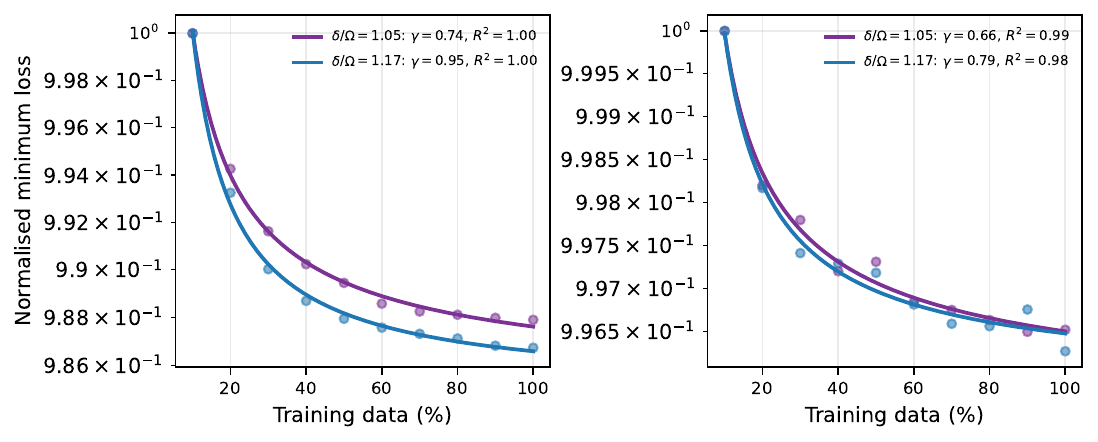}
  \caption{Normalised aggregate loss curves and fitted forms for the two detunings in Figure~\ref{fig:detuning-static-losses}. Points are observed seed means and lines are fits to Equation~\eqref{eq:data-scaling}. Each series is normalised by its largest observed loss for display only.}
  \label{fig:detuning-fit-curves}
\end{figure}

\subsection{Dynamics of the Fit Parameters}
\label{subsec:fit-parameter-dynamics}

Whilst the previous section considers data close to the critical point, a natural question is: how do the fitted parameters evolve as the detuning is varied away from criticality? To answer this question, this section analyses the induced flow of the data-scaling exponent $\gamma_D$, the goodness of fit of the power-law ansatz as quantified by $R^2$, and the asymptotic loss floor $\mathcal{L}_\infty$ across the low-detuning regime, through the critical point, and into the high-detuning regime. The values of $\delta$ are chosen such that they do not deviate \textit{too} far from $\delta=1.1$ and saturate the classical or resonant regimes.

\begin{figure}[H]
\centering
\includegraphics[width=0.98\linewidth]{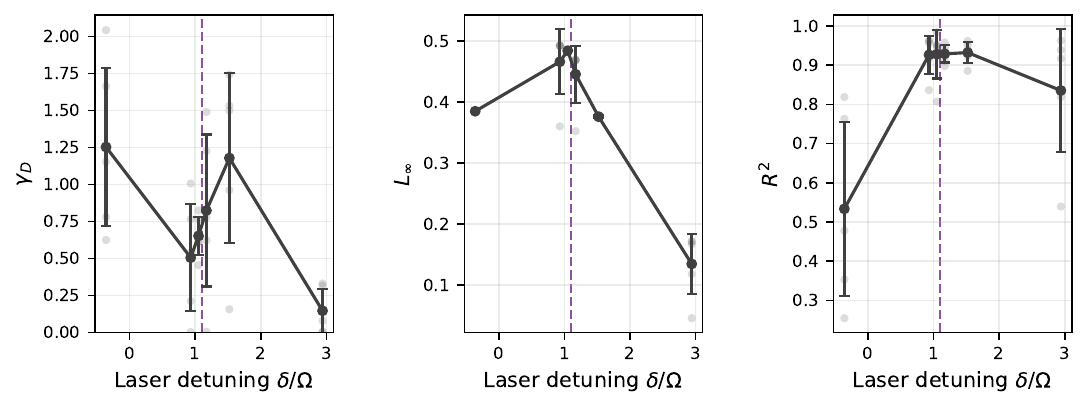}
\caption{Validation-fit parameters for the six retained detunings. Faint points denote fits to individual random seeds, while circles and error bars indicate the corresponding mean and standard deviation at each detuning. The dashed vertical line marks the critical point at $\delta/\Omega=1.10$.}
\label{fig:detuning-flow}
\end{figure}

Figure~\ref{fig:detuning-flow} extends the preceding analysis to six detuning values by displaying the results of validation fits performed independently for each random seed. The plotted spread directly characterises seed-to-seed variability in the fitted parameters and should not be interpreted as a confidence interval for a thermodynamic critical exponent. The values closest to the critical point at $\delta/\Omega=1.10$ are consistent with the aggregate results reported in Table~\ref{tab:detuning-data-scaling}.

The strongest agreement with the power-law ansatz is observed in the vicinity of the critical point. Near the dashed line in Figure~\ref{fig:detuning-flow}, the fitted $R^2$ values are both high and relatively tightly clustered, while the corresponding values of $\gamma_D$ remain of order unity. This behaviour is consistent with the aggregate fits in Table~\ref{tab:detuning-data-scaling}. At $\delta/\Omega=1.05$ and $1.17$, the training fits yield $R^2=0.998$ and $0.997$, respectively, while the corresponding validation fits yield $R^2=0.989$ and $0.983$. Likewise, the fitted curves in Figure~\ref{fig:detuning-fit-curves} closely track the normalised seed-averaged losses at these detunings.

By contrast, the fits become less stable towards the endpoints of the detuning sweep. In particular, at $\delta/\Omega=-0.36$ and $2.94$, the seed-level $R^2$ values are lower and exhibit a broader spread. The fitted exponents and asymptotic loss floors also vary more strongly between seeds. Although the Hoffmann form can still be fitted at an individual detuning far from the critical point, the resulting parameters are substantially less reproducible. 

% Figure~\ref{fig:detuning-flow} delineates the limits of this interpretation. The lower and more variable values of $R^2$ at the endpoints of the detuning sweep, together with the enhanced seed-to-seed variation in $\gamma_D$ and $\mathcal{L}_\infty$, indicate that the same fitting form is not stable far from the crossover. The Chinchilla ansatz is therefore used to characterise the near-critical data rather than as a single scaling law applicable uniformly across all detunings.

% \subsection{Near-Critical Behaviour}
% \label{subsec:nearcritical-behaviour}

As can be seen in Figure~\ref{fig:detuning-static-losses}, both the training and validation losses decrease smoothly with increasing dataset size at $\delta/\Omega=1.05$ and $1.17$. The fitted curves in Figure~\ref{fig:detuning-fit-curves} closely follow the corresponding normalised seed means, and all four aggregate fits reported in Table~\ref{tab:detuning-data-scaling} achieve high values of $R^2$ against the power-law ansatz. 

Taken together, these observations indicate that a power-law ansatz provides the most accurate empirical description of the sampled data near the critical point.
%, meaning that near-critical Rydberg models scale like natural language.

A possible explanation for the enhanced stability of this scaling form near the critical point is provided by the behaviour of the correlation length. At a continuous phase transition in the thermodynamic limit, the correlation length $\xi$ diverges, and the system develops fluctuations over arbitrarily large spatial scales. Although finite lattices and non-zero temperatures preclude a true divergence, one nevertheless expects $\xi$ to become large relative to its values away from the critical point; the resulting distribution contains correlations over a broad range of length scales, rather than being dominated by either short-range fluctuations or a single characteristic scale.
As the dataset size is increased, the model gains access to progressively better estimates of correlations at increasingly large separations and of higher-order configurations involving multiple spatial scales. 
Each additional increment of data presumably resolves further structure in the distribution, producing a gradual power-law improvement in the loss. 

Away from the critical point, where the correlation length is shorter, the relevant statistical structure may be exhausted more rapidly. In that regime, additional samples predominantly refine already-learned local statistics, and the simple power-law ansatz need not remain stable over the sampled range of dataset sizes as they are primarily informing the model about what it has already learnt.

These correlations provide additional learnable structure as the amount of training data is increased. The observed scaling form therefore suggests that Hoffmann-like behaviour may arise more generally from the organised and learnable structure of a probability distribution, rather than being specific to language. This analogy should nevertheless be interpreted cautiously: it does not imply that the fitted exponents are universal, nor that quantum-measurement data and natural language data share a common mechanism.

It is in this restricted sense that RydbergGPT near the critical point exhibits scaling behaviour analogous to that observed in natural language models. Natural language is organised at several levels that unfold over different lengths of a sequence. Nearby words are linked by grammar and local meaning, while longer spans of text carry sentence structure, and paragraphs communicate ideas. These patterns operate over a combination of short, medium, and large distances, since a word can depend not only on the words immediately before it, but also on information introduced much earlier. 
The fact that quantum models trained on near-critical data obey the same correspondence as language is non-trivial and somewhat remarkable.
%; quantum data is certainly not a priori expected to exhibit similar phenomenology to that of natural language!

\section{Mutual-information scaling for Rydberg and natural language data}
\label{sec:common-mi}

The scaling analysis in Section~\ref{sec:fixed_param_scaling_rydberg} shows that the fixed-model power-law ansatz is most stable near the critical point. We postulated that this observation could suggest that the scaling of the predictive loss may be connected to the organisation of the data distribution across varying spatial scales. Near the critical point, dependence persists over a broad range of separations, so additional samples can continue to refine statistical structure that is not confined to nearest-neighbour configurations. Natural language is also structured across several sequential scales. It is therefore useful to ask whether the two data sources possess quantitatively comparable ranges of statistical dependence, rather than relying only on an analogy between their loss curves.

Taking inspiration from \cite{Li1990MutualIF}, we use mutual information as a candidate observable for this comparison\footnote{A limitation of this approach is that it is sensitive to broken permutation symmetries within the Rydberg snapshot which the transformer is not.}. It detects arbitrary pairwise statistical dependence without assuming a linear relation, and it is defined for both binary occupations and categorical language symbols. It is also invariant under invertible re-labellings of a fixed alphabet. Its raw magnitude is not, however, directly comparable across the two domains. The local entropies are different, the Rydberg correlations are distributed over a two-dimensional lattice, and a language distance expressed in tokens changes with the tokenisation convention. 

Of course, these considerations inform the comparison used here. The same null-corrected and entropy-normalised metric is constructed in both systems; language is measured on a fixed character-level basis; and the resulting second-moment length is divided by the largest separation included in the measurement. 

\subsection{Definition of the common estimator}
\label{subsec:common-mi-estimator}

Let $X_i$ denote the discrete observable at location $i$. For the Rydberg data it is the binary occupation of a lattice site. For language it is the character-level symbol at a fixed position in the underlying text. The latter representation has a fixed alphabet and a fixed microscopic coordinate; it is consequently unchanged by any subsequent word or subword tokenisation.

For Rydberg sites at lattice coordinates $(x_i,y_i)$ and $(x_j,y_j)$, their separation in lattice-spacing units is
\[
 r_{ij}=\sqrt{(x_i-x_j)^2+(y_i-y_j)^2}.
\]
The displacement vector from site $i$ to site $j$ is the ordered coordinate difference $(\Delta x,\Delta y)=(x_j-x_i,y_j-y_i)$, so the two components give the signed horizontal and vertical offsets, respectively. The magnitude of the pair $(\Delta x, \Delta y)$ is given by $r_{ij}$. Let $\mathcal P_s$ denote the multiset of observed value pairs at separation $s$. For the Rydberg data, $\mathcal P_s$ contains $(X_i^{(q)},X_j^{(q)})$ for every configuration $q$ and every lattice pair with $r_{ij}=s$. For language, it contains $(X_t,X_{t+s})$ for every within-document position $t$ for which both symbols are present. Writing $N_s=|\mathcal P_s|$, the separation-conditioned empirical joint law and its marginals are
\begin{align*}
\widehat p_s(a,b)
&=\frac{1}{N_s}
\#\bigl\{(x,y)\in\mathcal P_s:x=a,\ y=b\bigr\},
\\
\widehat p_{s,L}(a)
&=\sum_b\widehat p_s(a,b),
\qquad
\widehat p_{s,R}(b)
=\sum_a\widehat p_s(a,b).
\end{align*}
Thus $s$ enters the estimator by selecting either a Pythagorean radial shell in the Rydberg array or a positional lag in the language sequence. The mutual information at that separation is
\begin{equation}
 \widehat I(s)
 =\sum_{a,b}\widehat p_s(a,b)
 \log\frac{\widehat p_s(a,b)}
 {\widehat p_{s,L}(a)\widehat p_{s,R}(b)}.
 \label{eq:common-pair-mi}
\end{equation}
\begin{samepage}
Mutual information estimated from a finite sample is generally non-zero even when the underlying variables are independent. This sampling bias is estimated by breaking the original pairing between observations. Here, a pair comprises two values observed at separation $s$. For the Rydberg data, $(X_i^{(q)},X_j^{(q)})$ contains the occupations of sites $i$ and $j$ in the same configuration $q$. For language, $(X_t,X_{t+s})$ contains the symbols at positions $t$ and $t+s$ in the same document. Breaking the pairing means retaining every observed value but changing which value at the first location is matched to which value at the second. For a fixed Rydberg site pair $(i,j)$, the occupations measured at site $i$ are left in their original order, whereas those measured at site $j$ are shuffled between configurations. Equivalently, $X_i^{(q)}$ is paired with $X_j^{(\pi(q))}$, where $\pi$ is a random permutation of the configuration labels; the two values therefore generally originate from different configurations. For language, the same operation is applied at each lag $s$: the first symbol in every observed pair is retained, while the second symbols are shuffled between pairs. Because only the ordering is changed, the occupation counts and the frequency of each language symbol are unchanged. The original co-occurrence pattern is nevertheless destroyed, so systematic dependence no longer contributes on average. Each shuffled data set is called a permutation-null sample. Its mutual information estimates the finite-sample contribution expected in the absence of dependence. Let $B$ denote the number of independent permutation-null samples generated at each separation, and let $\widehat I^{(m)}_{\rm null}(s)$ denote the mutual information estimated from sample $m\in\{1,\ldots,B\}$. The empirical marginal entropies are $\widehat H_L(s)=-\sum_a\widehat p_{s,L}(a)\log\widehat p_{s,L}(a)$ and $\widehat H_R(s)=-\sum_b\widehat p_{s,R}(b)\log\widehat p_{s,R}(b)$, with $0\log0$ defined to be zero.

Equation~\eqref{eq:common-normalised-mi} combines the two adjustments required for comparison across the Rydberg and language data. First, the mean permutation-null mutual information is subtracted from the observed value to remove the contribution expected from finite-sample bias. Second, the result is divided by the geometric mean of the two marginal entropies. Raw mutual information is measured in nats and its attainable magnitude depends on the uncertainty of the two variables; the entropy denominator removes this local scale. The geometric mean is symmetric under exchange of the two locations. Moreover, for exact distributions, $I(X;Y)\leq\min\{H(X),H(Y)\}\leq\sqrt{H(X)H(Y)}$, so the corresponding uncorrected entropy-normalised mutual information lies between zero and one. The resulting bias-corrected, dimensionless profile is
\begin{equation}
 \widehat\rho_{\MI}(s)
 =\frac{\widehat I(s)-B^{-1}\sum_{m=1}^{B}
 \widehat I^{(m)}_{\rm null}(s)}
 {\sqrt{\widehat H_L(s)\widehat H_R(s)}}.
 \label{eq:common-normalised-mi}
\end{equation}
\end{samepage}
For each language corpus, positional stationarity is assumed over the analysed prefix: the probability of observing a given symbol is taken not to depend on its absolute position within a document. The left and right marginal entropies in Equation~\eqref{eq:common-normalised-mi} are therefore estimated by the same corpus-wide symbol entropy. This assumption concerns only the single-symbol distribution; dependence between symbols at different positions remains the quantity being measured.

For the Rydberg data, the occupation probability, and hence the binary entropy, can differ between sites because the array has open boundaries. Equation~\eqref{eq:common-normalised-mi} is therefore evaluated separately for each site pair $(i,j)$: the null-subtracted mutual information for that pair is divided by $\sqrt{\widehat H_i\widehat H_j}$ before any spatial average is taken, where $\widehat H_i$ and $\widehat H_j$ are the empirical occupation entropies of the two sites. For a fixed displacement $(\Delta x,\Delta y)$, the resulting pair-normalised values are averaged over every origin for which both sites lie inside the array. These displacement averages are then combined for all displacements with the same Pythagorean length $s=\sqrt{\Delta x^2+\Delta y^2}$. Averaging in this order gives each displacement equal weight even when different numbers of its translations fit within the open boundary.

The population value of mutual information is non-negative, but the null-subtracted finite-sample estimate in Equation~\eqref{eq:common-normalised-mi} can fluctuate below zero. No monotonicity constraint is imposed: mutual information is not required to decrease at every successive separation, and shell-dependent or lag-dependent structure should remain in the estimator. Negative estimates are also not set to zero, since pointwise clipping would retain positive fluctuations while discarding comparable negative fluctuations and would therefore bias the range upwards.

The signed estimates are instead used directly in the cumulative zeroth and second moments. For a cutoff $R$, the second-moment length is
\begin{equation}
 \widehat\xi_{\MI,2}^{2}(R)
 =\frac{\displaystyle
 \sum_{0<s_k\leq R}g_d(s_k)s_k^2
 \widehat\rho_{\MI}(s_k)}
 {\displaystyle
 2d\sum_{0<s_k\leq R}g_d(s_k)
 \widehat\rho_{\MI}(s_k)}.
 \label{eq:common-second-moment}
\end{equation}
Here $R$ is the largest separation included in the sum, $d=2$ for the Rydberg array and $d=1$ for language, while $g_d(s)$ is the number of distinct displacement vectors, or signed coordinate offsets, whose magnitude is $s$. In one dimension the two directions supply a common factor which cancels. The factor $2d$ is the standard dimensional normalisation of a second-moment length. Appreciable dependence at large separations increases $\widehat\xi_{\MI,2}(R)$ through the factor $s_k^2$. The length is defined when the cumulative numerator and denominator are both positive; this condition holds throughout the cutoff ranges reported below.

% The resulting length is expressed in lattice spacings for the Rydberg data and in character positions for language. It is therefore divided by the same cutoff $R$ used to construct it:
% \begin{equation}
%  Q_{\MI}(R)=\frac{\widehat\xi_{\MI,2}(R)}{R}.
%  \label{eq:common-q}
% \end{equation}
% The ratio $Q_{\MI}(R)$ is dimensionless and measures the dependence range relative to the resolved observation window, rather than equating a lattice spacing with a character position. If dependence is confined to a finite microscopic range, $\widehat\xi_{\MI,2}(R)$ saturates and $Q_{\MI}(R)$ decreases as the cutoff is extended. An algebraic profile instead gives $\widehat\xi_{\MI,2}(R)\propto R$ over its scaling interval and hence an approximately constant $Q_{\MI}$.

\subsection{Data and estimation}
\label{subsec:common-mi-data}

The Rydberg calculation uses the open $6\times6$ occupation arrays at $R_b/a=1.15$, $\beta\Omega=8$ and the six detunings employed in the scaling analysis. Each profile is estimated from $10^5$ independent configurations, such that $R_{\max}=\sqrt{50}$ in lattice-spacing units (for a 6x6 array one has a maximum of 5 spaces in each direction, giving Pythagorean distance $\sqrt{50}$).

For language, the two common open natural language datasets AG News and Yahoo Answers are used. Each dataset is represented by a prefix containing 200k word-level symbols after a fixed text normalisation. Pairs are pooled only within documents, so no dependence is introduced across an artificial document boundary. Symbol lags $1\leq s\leq60$ are retained.

% Standard errors for $Q_{\MI}(R_{\max})$ are obtained from eight disjoint blocks: configuration blocks for the Rydberg ensembles and document blocks balanced by symbol count for the corpora.
% The covariance of the cumulative numerator and denominator in Equation~\eqref{eq:common-second-moment} is estimated across the blocks and propagated to $Q_{\MI}$ by the first-order delta method. This retains every signed block profile without requiring the moment ratio of each block to be positive.

\begin{figure}[H]
  \centering
  \includegraphics[width=0.98\linewidth]{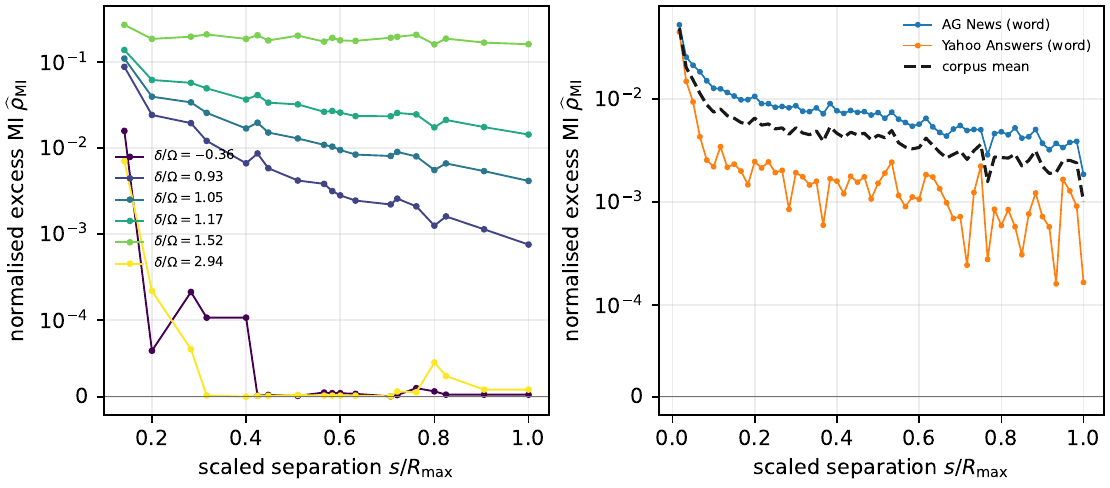}
  \caption{Entropy-normalised excess mutual-information profiles for the Rydberg occupations (left) and natural language symbols (right). Markers show the signed permutation-corrected estimates in Equation~\eqref{eq:common-normalised-mi}, and lines join successive separations. The symmetric-logarithmic vertical scale is linear for $|\widehat\rho_{\MI}|\leq10^{-4}$, so small negative finite-sample estimates remain visible. Separations are divided by the largest available cutoff within each system. The left panel contains the six open $6\times6$ Rydberg detunings. The right panel contains the AG News and Yahoo answers profiles in the fixed character-level representation; the dashed line is their pointwise arithmetic mean at each separation.}
  \label{fig:common-mi-profiles}
\end{figure}

Figure~\ref{fig:common-mi-profiles} shows that the detunings with the highest Hoffmann-fit $R^2$ values in Section~\ref{sec:fixed_param_scaling_rydberg} are also those whose Rydberg MI two-point functions are closest in shape to their natural language counterparts. The language profiles fall rapidly at short separation and then cross over to a shallower positive tail. The Rydberg curves at these high-$R^2$ detunings display a similar hierarchy: strong local dependence is accompanied by progressively weaker dependence over larger lattice distances. This differs from the remote configurations at $\delta/\Omega=-0.36$ and $2.94$, for which the corrected MI rapidly approaches the null level. For an autoregressive learner, a positive tail means that increasingly distant observations can still carry predictive information. Since this signal becomes weaker with separation, more samples are required to estimate it reliably; the model can therefore continue to learn structure at larger scales after the dominant local statistics have been resolved. By contrast, once the profile has collapsed to the null level, increasing the context range supplies little additional pairwise information. The numerical proximity of the corresponding dimensionless ranges should not be over-interpreted; what is important is the scaling behaviour. The coexistence of strong local information and a weaker extended tail during the critical point therefore provides supporting evidence for the hypothesis posed in Section~\ref{sec:fixed_param_scaling_rydberg}: the enhanced stability of Hoffmann-type scaling near the critical region is associated with learnable statistical dependence distributed across a hierarchy of scales.

\section{Conclusions}
\label{sec:conclusions}

With the advent of modern neural networks, neural scaling laws have become a crucial means of relating attainable loss to the number of available training tokens and model parameters. Their success in language modelling nevertheless leaves open a structural question: which properties of a training distribution permit a stable power-law learning curve? Do models trained on quantum data exhibit the same scaling behaviour as those trained on natural language? To address this, we tested fixed-model Hoffmann scaling across a detuning-controlled finite-size Rydberg critical point while holding the architecture, objective, and training procedure fixed. By comparing the goodness-of-fit of a power-law Hoffmann ansatz across several detunings, we assess whether quantum data exhibits language-like scaling across different physical regimes.

The central result is that, within an envelope surrounding the critical point, the loss is accurately described by a power-law ansatz consisting of an asymptotic loss floor and a power-law dependence on dataset size. The fitted curves attain high values of $R^2$ and exhibit limited seed-to-seed variation. At detunings far from the critical region, both fit quality and parameter stability are substantially reduced. The evidence therefore supports the Hoffmann form as an empirical description of near-critical data, but not as a detuning-independent scaling law for quantum data in general. More broadly, this suggests that stable power-law scaling is not determined by model architecture alone, but depends on the statistical regime of the training distribution.

Mutual-information analysis provides independent support for this interpretation. Rydberg configurations near the critical point show the closest agreement in functional form with natural language: both display strong short-range dependence followed by a decaying positive tail over a substantial fraction of the normalised observation range. By contrast, profiles at detunings distant from the critical point decay rapidly. Although this does not imply equality of correlation lengths or a universality-like equivalence between the two distributions, it indicates that the near-critical Rydberg data has the most language-like statistical organisation.

Taken together, the simultaneous occurrence of stable power-law fits and dependence distributed across several scales identifies multi-scale statistical structure as a plausible explanation for the remarkable validity of the scaling ansatz near the critical point. One interpretation is that such distributions continue to provide statistically useful structure as dataset size increases, allowing progressively smaller but systematic improvements in loss. Around the critical region, the correlation length is very large, meaning many samples are required to fully learn the distribution 
suggesting that scaling behaviour should be viewed as a property of the model–data pair, rather than of the model alone.

Several limitations constrain the scope of this conclusion. The mutual-information estimator depends on the construction of the finite-sample reference, and thus breaks the permutation and $\mathbb Z_2$ symmetry of the array. Quantum fluctuations, open boundaries, and the restricted lattice size also limit the physical interpretation,
however finite-size scaling in the underlying
Monte Carlo data are well-understood \cite{RGPTdata}.

%. One should nonetheless note that quantum Monte Carlo simulations indicate that finite-size effects in the training data are sub-leading, however. 

Nevertheless, given these results, we hypothesise that neural scaling laws may emerge most clearly for datasets whose statistical dependencies are distributed over multiple scales.
This suggests that for quantum simulators,  like Rydberg atom arrays, that are expected to continue producing larger and richer qubit measurement datasets in the future, robust neural scaling laws are most likely to be observed when the system is tuned near a quantum critical point.  More generally, our work has provided a possible link
between the structure of a training distribution and the form of its learning curve.

\section*{Data Availability Statement}

The model used in this work is based on \cite{Fitzek_2025}. The modified model is available at \url{https://github.com/PIQuIL/Quantum-Chinchilla}.

\section*{Acknowledgements}
\label{sec:acknowledgements}
We thank Yi-Hong Teoh and Schuyler Moss for crucial discussions on the RydbergGPT model.
DSB and AGS are grateful to 
Pierre Andurand for his donation supporting this research. DSB is partially supported by the Science and Technology Facilities Council (STFC) Consolidated Grant ST/X00063X/1 ``Amplitudes, Strings \& Duality''.  
RGM would like to acknowledge the support of the Natural Sciences and Engineering Research Council of Canada (NSERC).
This research was also supported in part by grant NSF PHY-2309135 to the Kavli Institute for Theoretical Physics (KITP). 
Research at the Perimeter Institute is supported in part by the Government of Canada through the Department of Innovation, Science and Economic Development Canada and by the Province of Ontario through the Ministry of Economic Development, Job Creation and Trade. YJK acknowlegdes the support from the National Science and Technology Council of Taiwan (NSTC) through grants 113-2112-M-002-033-MY3 and  115-2124-M-001-015. We acknowledge the National Center for High-Performance Computing in Taiwan for the computational resources used in this research.

\appendix

\section{Mutual-information dependence in Rydberg samples}
\label{sec:rydberg-mi}

For occupations at sites $i$ and $j$, we estimate the plug-in pairwise mutual information
\begin{equation}
\widehat I_{ij}=
\sum_{a,b\in\{0,1\}}\widehat p_{ij}(a,b)
\ln\frac{\widehat p_{ij}(a,b)}
{\widehat p_i(a)\widehat p_j(b)},
\label{eq:pair-mi}
\end{equation}
where zero-probability terms are omitted.  For sites at lattice coordinates $(x_i,y_i)$ and $(x_j,y_j)$, we define the radial separation in lattice-spacing units by the Pythagorean distance
\begin{equation}
r_{ij}=\sqrt{(x_i-x_j)^2+(y_i-y_j)^2}.
\label{eq:lattice-distance}
\end{equation}
We average the pairwise values over all open-boundary pairs at a common separation $r_{ij}=r$ to obtain $\widehat I(r)$ for the same six detunings presented in Section~\ref{sec:fixed_param_scaling_rydberg}, using 10,000 configurations sourced from Quantum Monte Carlo data.

\begin{figure}[t]
\includegraphics[width=0.6\linewidth]{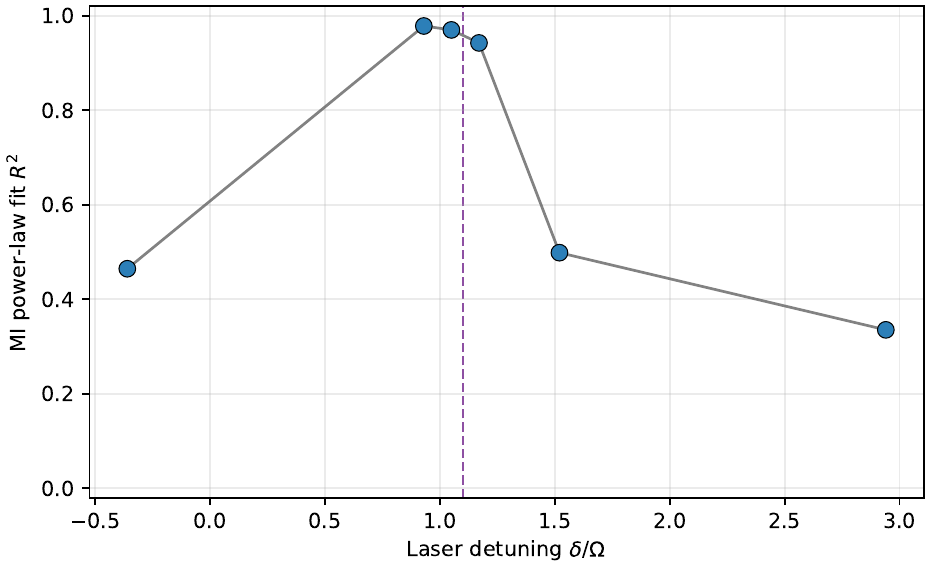}

    \caption{Fit quality of the mutual-information power-law ansatz over detuning.  Points give the $R^2$ of the log--log fit in Equation~\eqref{eq:mi-power-law}; the dashed line marks the critical point.}
    \label{fig:detuning-mi-summary}
\end{figure}

At criticality, the mutual-information scaling ansatz has the algebraic form below, with $\eta$ the mutual-information scaling exponent.  As a finite-array diagnostic, we fit the 19 radial shells in the interval $1\leq r\leq\sqrt{50}$ to
\begin{equation}
I(r)=A r^{-\eta}.
\label{eq:mi-power-law}
\end{equation}
Thus $\log I(r)=\log A-\eta\log r$ is fitted by ordinary least squares over the positive radial-shell estimates.  In the near-critical reference region, the high fit qualities in Figure~\ref{fig:detuning-mi-summary} show that this ansatz describes the measured radial dependence well.  The fitted $\eta$ is nevertheless a descriptive finite-array parameter; it is not identified with a universal critical exponent.

The fitted exponent decreases from $\eta=2.24$ at $\delta/\Omega=0.93$ ($R^2=0.978$) to $\eta=1.56$ at $1.05$ ($R^2=0.970$) and $\eta=1.08$ at $1.17$ ($R^2=0.942$).  Within the adopted ansatz, this corresponds to a progressively slower radial decay on approaching and passing the critical point.  The values $R^2=0.978$, $0.970$ and $0.942$ at $\delta/\Omega=0.93$, $1.05$ and $1.17$, respectively, provide direct support for the use of the mutual information ansatz as a proxy for correlation length.  At the more remote settings, the fit is less stable: $\eta=1.57$ at $\delta/\Omega=-0.36$ ($R^2=0.465$), $\eta=0.163$ at $1.52$ ($R^2=0.499$), and $\eta=1.71$ at $2.94$ ($R^2=0.335$).  The power-law parameters therefore provide a compact description of the measured radial dependence, but do not establish a universal exponent or a sharply defined phase transition.  This limitation is physically expected for a finite, thermally mixed quantum system.

\bibliographystyle{JHEP}
\bibliography{bib}
\end{document}